\documentclass[conference]{IEEEtran}
\IEEEoverridecommandlockouts
\usepackage{cite}
\usepackage{amsmath,amssymb,amsfonts}
\usepackage{algorithmic}
\usepackage{graphicx}
\usepackage{textcomp}
\usepackage{xcolor}
\usepackage{booktabs}
\usepackage{array}
\usepackage{enumitem}
\usepackage{multirow}
\usepackage{url}

\newcommand{\Cid}[1]{C#1}
\newcommand{\Lid}[1]{L#1}

\def\BibTeX{{\rm B\kern-.05em{\sc i\kern-.025em b}\kern-.08em
    T\kern-.1667em\lower.7ex\hbox{E}\kern-.125emX}}
\begin{document}

\title{A-CPES: A Reference Framework for Agentic AI\\in Cyber-Physical Energy Systems
\thanks{}
}

\author{\IEEEauthorblockN{1\textsuperscript{st} Xiaoyu Zhang}
\IEEEauthorblockA{\textit{School of Information Science and Engineering} \\
\textit{Northeastern University}\\
Shenyang, China \\
2410400@stu.neu.edu.cn}
\and
\IEEEauthorblockN{2\textsuperscript{nd} Qiuye Sun}
\IEEEauthorblockA{\textit{State Key Laboratory of Synthetical Automation for Process Industries} \\
\textit{Northeastern University}\\
Shenyang, China \\
sunqiuye@ise.neu.edu.cn}
\and
\IEEEauthorblockN{3\textsuperscript{rd} Jiachen Xu}
\IEEEauthorblockA{\textit{Department of Computer Science} \\
\textit{Aalborg University}\\
Aalborg, Denmark \\
jiachenxu@cs.aau.dk}
\and
\IEEEauthorblockN{4\textsuperscript{th} Zhongming Yao}
\IEEEauthorblockA{\textit{College of Computer Science and Technology} \\
\textit{Zhejiang University}\\
Hangzhou, China \\
yaozzzm@gmail.com }
\and
\IEEEauthorblockN{5\textsuperscript{th} Yushuai Li}
\IEEEauthorblockA{\textit{Department of Computer Science} \\
\textit{Aalborg University}\\
Aalborg, Denmark \\
yusli@cs.aau.dk}
}

\maketitle

\begin{abstract}
Energy system operation contains a loop of work that automation has never taken over: posing the optimization problem the current cycle should solve, disposing of infeasibility, sequencing a solution into interlocked switching orders, assembling evidence no single model holds, negotiating adjustable capacity with many parties, and settling experience into practice. Licensed dispatchers carry all of it in person, and the rising share of variable renewable generation is making that loop turn faster than their number can grow. Agentic AI supplies the abilities it requires, but enters as the outer loop of control: it calls SCED and the other decision models rather than being called by them. We propose A-CPES, three nested rings, an authorization and accountability frame around an agentic control outer loop around a six-layer CPES core. We argue the loop is indivisible, tune where and how tightly it may close, state eight structural failure modes as falsifiable predictions, and specify six governance modules that rebuild the authorization frame until it covers the loop, before the loop starts turning.
\end{abstract}


% =========================================================================
\section{Introduction}

\begin{figure*}[!t]
\centering
\includegraphics[width=181mm, trim=4 4 4 4, clip]{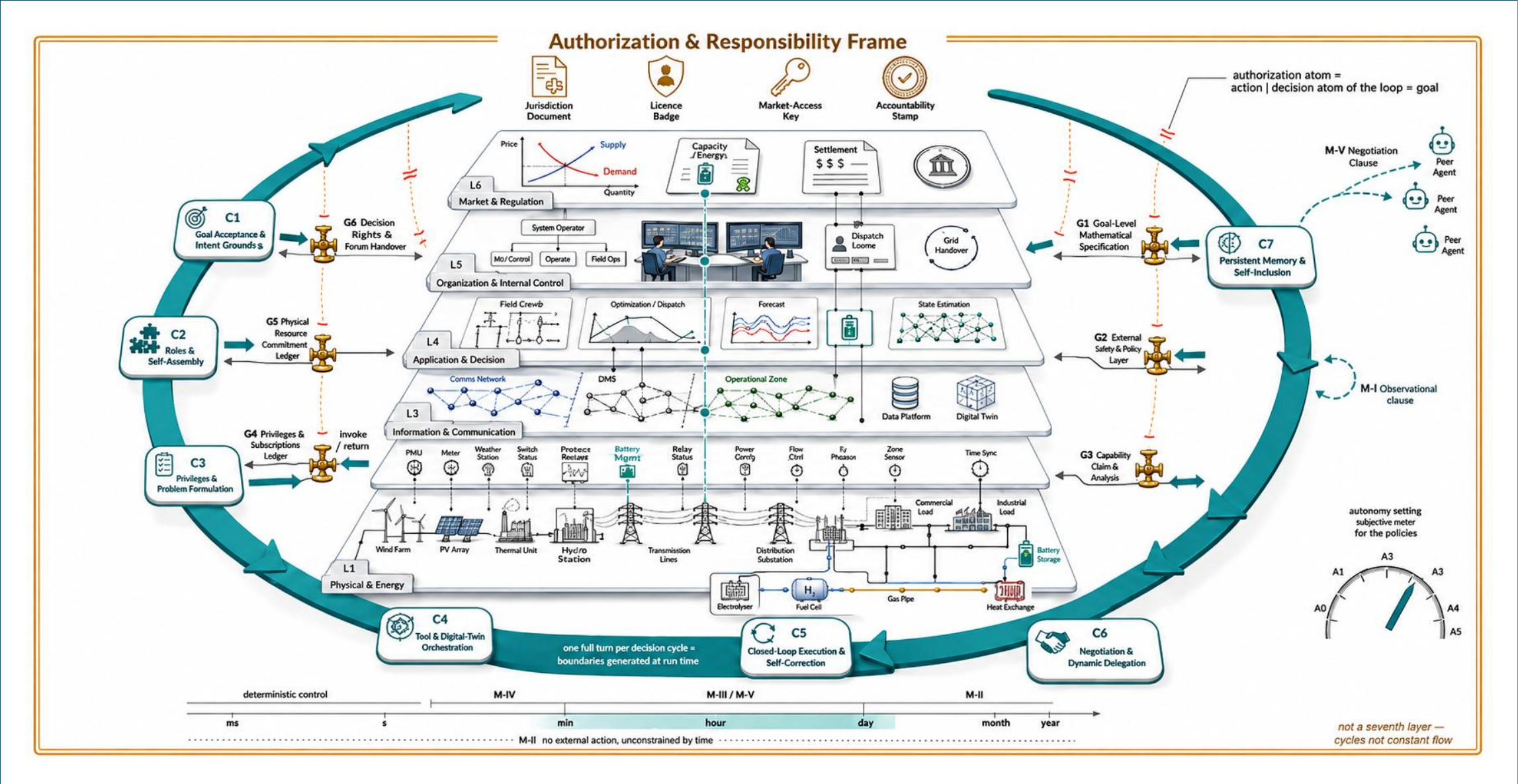}
\caption{The A-CPES framework. Every boundary in the six-layer core is drawn ex ante and statically. The outer loop turns one full revolution per decision cycle, traversing all six channels, and it is the initiator: it calls SCED and the other models, not the reverse. The outermost ring is the authorization frame, not a seventh layer: it takes no part in control flow and reaches the channels only through the gates G1--G6. The broken dashed lines from frame to gates are the core diagnosis: the frame's atom of authorization is the action while the loop's is the goal, so the frame does not cover the loop today.}
\label{fig:acpes}
\end{figure*}

Energy system~\cite{Li} operation contains a loop of work that automation has never taken over: deciding which optimization problem the current cycle should pose, diagnosing and disposing of infeasibility, translating a solution into switching orders that carry sequence and interlocks, assembling information no single model holds, negotiating adjustable capacity with many external parties, and settling what worked into standing practice. Licensed dispatchers carry all of it in person.

Rising variable renewable generation~\cite{YuPV} makes that loop turn faster and more often: intra-day rolling rescheduling multiplies, uncertainty widens, infeasible cases become routine. Distributed resources and aggregators~\cite{Xu} multiply the parties to coordinate; multi-energy coupling~\cite{YuReward} multiplies the adjustment options and their combinations. The loop's workload grows at the speed of system scale, whereas the number of licensed dispatchers cannot grow in proportion.

Existing automation attaches to the system at points: model predictive control~\cite{ref:mpc}, reinforcement learning~\cite{YuSurvey} and large language models~\cite{ref:llm4power} each do one designated thing, and none initiates, plans or closes a loop. Agentic AI can take this loop whole and keep it turning. It accepts a goal rather than an instruction, breaks it into steps itself, decides what to read and which tool to invoke~\cite{ref:react,ref:toolformer}, executes closed loop and corrects itself against the result~\cite{ref:reflexion}, delegates to other agents~\cite{ref:autogen}, and carries experience across sessions~\cite{ref:voyager}. These are exactly the abilities the loop's segments require, so supply and demand are meeting. Yet such systems remain rare in energy, and the few prototypes that exist~\cite{ref:gridmind,ref:gridagent,ref:xgridagent} close their loop on a simulator.

Operators preparing to deploy are stuck on three questions. Which segments do we hand over, and where does a human take back? No vocabulary today says what the loop is made of (\S\ref{sec:loop}, \S\ref{sec:tune}). Can we hand over only part of it? Starting small feels safest, but nobody has checked how much value survives (\S\ref{sec:indiv}). How much autonomy, on what evidence may it be raised, and how is an incident traced? Existing scales are product tiers with no notion of setting or revalidating a level, and no agent counterpart of the sequence-of-events record (\S\ref{sec:auto}, \S\ref{sec:gov}).

The root cause is that no single picture holds both languages at once: the authorization engineering of energy systems (mandate division, change-controlled settings, work permits, type testing, fault recording) and the agent capability language of AI (tool use, planning, memory, delegation). Both are mature, neither maps onto the other, and none of the three questions can be posed completely in either. Authorization here is moreover never granted after the fact, type testing and work permits putting the rule before the equipment, so writing the specification only once the loop is already turning would let an accomplished fact force the rule.

What is missing is therefore not a stronger model, nor another simulator-bound system, but a reference picture that aligns the two languages before deployment. We propose A-CPES, three nested rings (Fig.~\ref{fig:acpes}, detailed in \S\ref{sec:frame}): an authorization and accountability frame enclosing an agentic control outer loop, which encloses a six-layer CPES core. With it come a tuning of where the loop closes, on what time scale and at what autonomy (\S\ref{sec:tune}), an atlas of eight failure modes (\S\ref{sec:fail}), and an enabling substrate with six governance modules (\S\ref{sec:gov}). It answers the three questions and establishes that the outer loop is indivisible: the smallest unit of coupling is not an integration point but a whole loop.

\section{The A-CPES Framework}\label{sec:frame}

A-CPES renders the coupling as three nested rings (Fig.~\ref{fig:acpes}): a six-layer CPES core, the callee in control flow; an agentic control outer loop, the initiator, whose reach, tool set and delegation chain are generated at run time; and an authorization and accountability frame, fixing whose action this is and who answers for it.

\textit{Inner core.} The core is anchored in SGAM, IEC 62357 and the NIST Smart Grid Framework~\cite{ref:sgam,ref:iec62357,ref:nist} rather than invented here: the six layers re-cut SGAM's domain-zone structure along gatekeeping mechanism, and Fig.~\ref{fig:acpes} carries their contents. What matters here is when each boundary is drawn. \Lid{5} and \Lid{6} are bounded ex ante and institutionally, by dispatch jurisdiction, mandate documents and certification, and by market admission and settlement identity. \Lid{2} and \Lid{4} are bounded at design time, by measurement redundancy and time synchronization~\cite{L5,L7}, and by model scope, interfaces and version-frozen study cases; \Lid{4} hosts SCED~\cite{ref:scopf} and state estimation~\cite{ref:se}. \Lid{3} is bounded ex ante and mandatorily, by enforced isolation between the production-control and management-information zones. \Lid{1} is bounded by physical law, the N-1 criterion and device settings, and multi-energy coupling sits there rather than in a layer of its own, because a coupling asset is one physical object carrying commitments to several markets at once (P6). All six boundaries are therefore drawn ex ante and statically, which suffices for conventional automation, since it works inside a layer and crosses between layers over predefined interfaces.

\textit{Middle ring.} The loop's seven segments (\Cid{1} accept goal, \Cid{2} self-assemble context, \Cid{3} pose the problem, \Cid{4} orchestrate tools, \Cid{5} execute, \Cid{6} delegate, \Cid{7} consolidate) land on different layers, so every revolution traverses all six. The loop is the initiator, calling SCED and the other models rather than the reverse; and its boundaries are generated at run time, in what it reads, which tools it combines and whom it delegates to. Hence static boundaries cannot contain a dynamically generated loop.

\textit{Outer frame.} The frame is not a seventh layer: it takes no part in control flow, and only decides whether the loop's traversals are permitted. Its atom of authorization is the action, an order, an operating step or a setting value, drawn ex ante from the same source as the layers' boundaries. So once the loop turns a full revolution, the same agent sits in two opposite relations: in control it is outside the energy system, in authorization it must remain inside it. \S\ref{sec:fail} derives eight structural consequences of this misalignment; \S\ref{sec:gov} rebuilds the frame until it covers.

% =========================================================================
\section{The Agentic Outer Loop}\label{sec:loop}
% =========================================================================

% Table II declared here, not in Sec. V: a table* can only be placed at a
% page top at or after its declaration point, so declaring it in Sec. V
% strands it on the last page.
% Table 1 -- failure-mode atlas (spans both columns, page top)
\begin{table*}[t]
\caption{Failure-mode atlas. Inclusion test: a phenomenon that still holds under
``RL $+$ frozen weights $+$ a projection layer'' is not listed, so the last
column being uniformly ``no / partly'' is deliberate. Rows follow outer-loop
segment \Cid{1}$\rightarrow$\Cid{7}.}
\label{tab:fail}
\centering
\scriptsize
\setlength{\tabcolsep}{4pt}
\renewcommand{\arraystretch}{1.0}
\begin{tabular}{@{}l>{\raggedright\arraybackslash}p{6.1cm}cccc>{\raggedright\arraybackslash}p{4.6cm}@{}}
\toprule
\textbf{\#} & \textbf{Failure mode} & \textbf{Segment} & \textbf{Layers} &
\textbf{Module} & \textbf{Property} & \textbf{Does it hold under RL?} \\
\midrule
P1 & Context self-assembly crosses the mandatory security-zone boundary, and is
the more offending the more useful & \Cid{2} & \Lid{3} & G4 & Security &
No \textbar{} the observation vector is fixed \\

P2 & Natural-language evidence becomes a new control channel, so prompt
injection~\cite{ref:injection} becomes a question of where an order came from
& \Cid{2} & \Lid{3} & G4 & Security &
No \textbar{} the observation vector is strongly typed \\

P3\,$\bigstar$ & Relaxing a constraint when the problem is infeasible $=$
exercising an authority never granted & \Cid{3} &
\Lid{4}$\rightarrow$\Lid{1}/\Lid{5} & G1, G2 & Safety &
No \textbar{} changing the problem is not in the action space \\

P4 & Digital-twin fidelity gap, with the twin selected by the agent at run
time & \Cid{4} & \Lid{3}, \Lid{4} & G3, E & Robustness &
Partly \textbar{} run-time self-selection of fidelity is agentic-specific \\

P5 & Terminal state satisfies N-1 while an intermediate state violates limits &
\Cid{5} & \Lid{1} & G1 & Safety &
No \textbar{} intermediate time steps are explicit decision variables \\

P6\,$\bigstar$ & Multi-level dynamic delegation double-promises one physical
resource, so the reserve on the books is fictitious & \Cid{6} &
\Lid{5}, \Lid{6} & G5 & Accountability & No \textbar{} there is no run-time
delegation \\

P7 & Memory forms operating practice that was never approved, and never raises
an alarm & \Cid{7} & \Lid{5} & G6 & Auditability &
No \textbar{} behavior stops drifting once weights are frozen \\

P8 & Autonomy is earned under normal conditions but must be redeemed under rare
ones & \Cid{7} & cross-cut & G3 & Robustness &
No \textbar{} the object under test is stable \\
\bottomrule
\end{tabular}
\end{table*}

Solvers, model predictive control and reinforcement learning all authorize objects that can be frozen: a constraint set and an objective, a model and a cost, an action space and frozen weights. Such objects are reviewable before deployment, exhaustively testable and version-sealed, which is why type testing and change-controlled settings apply to them. They are equipment, not people. The agentic loop offers no such object: what it does this revolution depends on what it read (\Cid{2}), what it remembers (\Cid{7}), which tools it holds (\Cid{4}) and whom it delegated to (\Cid{6}).

\subsection{The seven segments}

\Cid{1} goal grounding attaches an operating goal stated in natural language to procedure clauses and market rules. \Cid{2} context self-assembly joins measurements, asset records, outage plans, weather and market information on its own initiative, and is the core increment an agent brings. \Cid{3} planning and problem posing decides what to pose and what to do when nothing is feasible. \Cid{4} tool and digital-twin orchestration combines solvers, simulators, forecasters and clearing engines at run time. \Cid{5} closed-loop execution and self-correction turns a solution into switching operations that carry sequence. \Cid{6} negotiation and dynamic delegation assembles adjustable resources across dispatch levels, aggregators and users. \Cid{7} persistent memory accumulates experience across sessions and forms operating preferences, and what settles there is organizational knowledge. Which layer each touches is drawn as the six channels of Fig.~\ref{fig:acpes}; which failure modes each produces is the ``segment'' column of Table~\ref{tab:fail}.

\subsection{Indivisibility}\label{sec:indiv}

Could one adopt only some, plugging the agent in where needed? Without \Cid{1} a human must still translate the goal into a formal specification, the most labor-intensive step to begin with. Without \Cid{2} it is a script calling a solver over study cases already in \Lid{4}, while what an infeasible case needs, outage reality, a weather swing, a market signal, is not in \Lid{4}. Without \Cid{3} the assembled evidence has nowhere to be used and the agent is an expensive dashboard. Without \Cid{4} it talks but does not compute, so its output is not engineering-acceptable. Without \Cid{5} it stops at a plan rather than a command, leaving sequence, interlocks and procedural safety on the human. Without \Cid{6} the problems it poses have nobody to fulfill them. Without \Cid{7} it starts from zero each revolution, though the value of rolling dispatch lies exactly across cycles.

Incomplete agentic is ineffective agentic: the seven are pairwise preconditions forming a closed loop, so it must be given whole, and given whole it necessarily traverses all six layers. That is a structural necessity rather than an implementation choice.

% =========================================================================
\section{Tuning the Outer Loop}\label{sec:tune}
% =========================================================================

The loop must be given whole, so one question remains: where does it close, on what time scale, and how tightly, three graduations of one dial.

\subsection{Where the loop closes: five modes}

In all five modes, all seven segments are running; what differs is where the loop closes. The human is not a removed segment but a gate on the loop. The independent variable is the closing layer, where the loop's own action lands after one revolution with no human in between. Candidates are ``nowhere'' plus the six layers, but \Lid{2} and \Lid{3} cannot receive, since closing means the output becomes a commitment or an action there, whereas a measurement and a communication channel are only read and passed through. That leaves $\{\emptyset,\Lid{5},\Lid{4},\Lid{1},\Lid{6}\}$, and the closing layer decides everything else. On $\emptyset$ there is no external action and no time-scale constraint: M-I, observation closure. On \Lid{5} the human sits on the gate and every revolution passes it, so the scale follows the human decision cycle into hours to days: M-II, advisory closure (day-ahead plans, outage scheduling). On \Lid{4} problems and plans carry checkpoints naturally, so the human samples by segment at minutes to hours: M-III, approval closure, the running example here being intra-day rolling dispatch. On \Lid{1} there is no time for segment boundaries, so the human samples by time and holds an envelope, escalating on excursion, at seconds to minutes: M-IV, envelope closure (AVC, storage scheduling). On \Lid{6} a signed commitment cannot be withdrawn, leaving only after-the-fact audit on the settlement cycle: M-V, negotiation closure (VPP aggregation).

M-I--M-IV descend the stack monotonically while the gate retreats from ``every revolution'' to ``sampled by segment'' to ``holds an envelope only'', whereas M-V goes outward to other parties, closing on commitments rather than on a physical quantity, which is why it alone owns P6 and G5.

\subsection{Time scale and choice of closing position}

The loop does not close onto the millisecond-to-second protection and primary frequency control loops, where determinism and type-testability are required and its run-time variability is a negative asset. Its value band is minutes to days, exactly where a human is already deciding on cross-layer information. In Fig.~\ref{fig:acpes}, M-I spans the whole axis because it does not close outward.

Three criteria choose the position: how much cross-layer information the decision needs, how physically irreversible an error is, and how much reaction time the human has. Intra-day rolling dispatch lands on M-III: information is large, errors are partly correctable next cycle, humans have minutes, and M-IV would be excessive because disposing of an infeasible case touches whose authority is whose. M-II is insufficient, not because the agent does fewer segments but because the gate sits on the earliest stretch, so every revolution's output lands on a human who must digest it and act within a fifteen-minute cycle. The gate cannot open and close as fast as the loop turns, which is the supply-demand gap of \S1 cashed out.

\subsection{Autonomy levels A0--A5}\label{sec:auto}

Closing position says where the loop closes; autonomy says how tightly. A is the dial, M is where the loop closes when the dial is at that notch, as the right-hand ruler of Fig.~\ref{fig:acpes} is drawn. Table~\ref{tab:auto} is a setting schedule; it lists no time scale, that being the intrinsic rhythm of the closing layer.

\begin{table}[t]
\caption{Autonomy setting schedule. A0 (no adoption) omitted.}
\label{tab:auto}
\centering
\scriptsize
\setlength{\tabcolsep}{3pt}
\renewcommand{\arraystretch}{0.98}
\begin{tabular}{@{}>{\raggedright\arraybackslash}p{0.70cm}>{\raggedright\arraybackslash}p{2.32cm}>{\raggedright\arraybackslash}p{0.88cm}>{\raggedright\arraybackslash}p{2.88cm}@{}}
\toprule
\textbf{Level} & \textbf{Authorized object (width of the G1 spec)} &
\textbf{Modules} & \textbf{Evidence required to promote} \\
\midrule
A1 read-only & reach of retrieval & G4, G6 & provenance labels complete, replayable \\
A2 advisory & $+$ admissible problem family & $+$ G1 & accepted/rejected advice is explicable \\
A3 bounded orch. & $+$ admissible relaxation set (pose, not execute) & $+$ G2, G3 & under rare conditions, relaxation choices track physical importance \\
A4 bounded loop & $+$ trajectory envelope $+$ commitment quota & $+$ G5 & whole-trajectory compliance (not terminal state) $+$ zero commitment conflicts \\
A5 supervised & the full quadruple & all $+$ drills & takeover-drill records across rare conditions \\
\bottomrule
\end{tabular}
\end{table}

Autonomy is not a product tier but a quantity to be set like a protection setting, and revalidated whenever memory or the tool set changes, since otherwise the thing under test is no longer the thing that was tested (P8), and promotion evidence must be drawn from rare operating conditions rather than accumulated running hours. A separate condition governs continued possession: A4 and A5 require a periodic unplug-the-agent drill, withdrawing the agent and measuring whether on-duty staff can take over within the original time scale. This mirrors the anti-accident drill regime, with the object under test changed from equipment failure to agent absence, and is necessary because uneventful A4/A5 operation lets takeover skill decay silently~\cite{ref:bainbridge}, a decay that appears in none of the agent's own metrics.

% =========================================================================
\section{Failure Modes and Governance Modules}\label{sec:diag}
% =========================================================================

\subsection{The failure-mode atlas}\label{sec:fail}

Table~\ref{tab:fail}'s eight entries are necessary consequences of one structure, a complete outer loop turning on six statically bounded layers, stated as falsifiable predictions because such loops today close on simulators. Its last column answers ``isn't this just safe RL?'': an RL violation is detectable because the problem did not change; an agentic relaxation is not, because the changed problem became the baseline the check is run against (P3).

\subsection{Enabling substrate and governance modules}\label{sec:gov}

The six modules below are not shackles on the agent: each rewrites one stretch of the authorization frame from ``per action'' to ``per goal and envelope'', until the frame covers the loop. Each names one reviewable, version-controlled object and pins to it something the loop generates at run time, and every isomorphism below runs from energy to AI, not the reverse. Three of them gate the channels an enabling substrate must open first, for \Cid{2} retrieval, \Cid{4} tool use and \Cid{6} delegation, only \Cid{5} already having one.

\begin{description}[leftmargin=0pt,itemsep=0.5pt,topsep=2pt,parsep=0pt]
\item[G1 Goal-level authorization specification] rewrites what is authorized:
orders give way to $\langle$ problem family; relaxation set; trajectory
envelope; commitment quota $\rangle$~\cite{L3,L6}. The freezable object agentic systems lack;
cf.\ graded dispatch authority. (P3, P5)

\item[G2 External safety veto layer] rewrites where safety is imposed: the projection applies to the problem, not the action, since overreach happens at problem posing; shielding~\cite{ref:shielding,ref:cpo} raised one level. (P3)

\item[G3 Capability-closure analysis] rewrites the unit of competence: the tool combination's closure is authorized, not the tool list; cf.\ recomputing settings after a primary-circuit change. (P4, P8)

\item[G4 Provenance and security-zone labeling] rewrites the admission of evidence: source and zone labels, cross-zone use refused, natural language graded apart from measurements; cf.\ secondary-system security rules. (P1, P2)

\item[G5 Physical resource commitment ledger] rewrites the bookkeeping of commitments: each resolves to a unique redeemable physical quantity, globally exclusive across carriers; cf.\ reserve-capacity registers. (P6)

\item[G6 Decision recording and human handover] rewrites record and handover:
freeze the input snapshot, retrieval hits, model version and seed, because an
agent has no SOE; cf.\ fault recording and shift handover. (P7)
\end{description}

% =========================================================================
\section{Conclusion}
% =========================================================================

Coupling agentic AI with energy systems is not a matter of adding AI at a few points, but of letting one autonomous outer loop wrap the entire stack and turn. At that moment containment holds in control flow, while the frame of authorization and accountability still sits in the old per-action, per-layer form. The engineering to be done is to rebuild that frame until it covers, before the loop starts turning.

% =========================================================================
{\scriptsize
\setlength{\itemsep}{0pt}
}

\end{document}